# Deep Learning-Driven Approach for Handwritten Chinese Character Classification

Kriuk Boris[a,b,*], Kriuk Fedor[a,b]

[a]Hong Kong University of Science and Technology, School of Engineering, Department of Computer and Electrical Engineering, Clear Water Bay, Kowloon, Hong Kong SAR, 999077

[b]Sparcus Technologies Limited, 50 Stanley Street, Central, Hong Kong SAR, 999077

Abstract. Handwritten character recognition (HCR) is a challenging problem for machine learning researchers. Unlike printed text data, handwritten character datasets have more variation due to human-introduced bias. With numerous unique character classes present, some data, such as Logographic Scripts or Sino-Korean character sequences, bring new complications to the HCR problem. The classification task on such datasets requires the model to learn high-complexity details of the images that share similar features. With recent advances in computational resource availability and further computer vision theory development, some research teams have effectively addressed the arising challenges. Although known for achieving high accuracy while keeping the number of parameters small, many common approaches are still not generalizable and use dataset-specific solutions to achieve better results. Due to complex structure, existing methods frequently prevent the solutions from gaining popularity. This paper proposes a highly scalable approach for detailed character image classification by introducing the model architecture, data preprocessing steps, and testing design instructions. We also perform experiments to compare the performance of our method with that of existing ones to show the improvements achieved.

## 1 Introduction

East Asian scripts may contain more than 6000 unique characters per work, making the classification task highly detail-dependent. In machine learning, the task of classifying detailed images that share multiple similar features is called fine-grained image classification.

Although convolutional neural networks[1,2] have forever revolutionized deep learning, the scientific community has yet to be satisfied with the results achieved on numerous computer vision problems, continuing to push the progress further. One of the remaining challenges is fine-grained image classification with a combination of several important complications that should be addressed together:

1. High Dimensionality: Detailed images, especially high-resolution ones, contain a vast amount of information. This can lead to the "curse of dimensionality," where the feature space becomes so large that it is difficult for classification algorithms to operate efficiently and effectively.

2. Imbalanced Dataset: In many real-world datasets, some classes are underrepresented. This can cause classifiers to become biased towards the more common classes and perform poorly on the rare ones.

3. Background Complexity: The complexity of backgrounds and the context in detailed images can confuse classification models, leading to misclassifications if the background is mistaken for the object of interest or vice versa.

4. Intra-class Variation: Detailed images within the same category can have a wide range of appearances, poses, lighting conditions, and occlusions. This variation makes it challenging to build models that generalize well across the entire category.

5. Computational Resources: Processing and classifying detailed images require significant computational power and memory, which can be a limitation for production.

Many approaches to tackle the following challenges have been made in recent years,[3–6] resulting in a better understanding of the individual fundamental steps needed to achieve higher model performance on fine-grained image classification. Some of the ideas, such as feature extraction, data augmentation, attention mechanisms, and ensembling, were also successfully implemented on the HCR datasets.[7]

In recent years, the machine learning community has made significant progress in understanding convolutional neural networks (CNNs) and the process of creating larger models[2,8,9] efficiently. However, the task of scaling CNNs remains a complex challenge, and researchers cannot solely rely on the established ideas that were initially proposed for unifying the theory of the scaling process. Even modern deep convolutional neural networks, designed to handle overfitting, face learning plateaus with time and stop learning important features after baseline, meaning that a task-specific approach is required to improve the results further. This creates a barrier for the deep learning models to extract continuously more complex features that are crucial for achieving higher precision on fine-grained image classification. The following challenge often prevents research teams from achieving the desired results in the character classification stage after choosing segmentation-based algorithms for text recognition, mainly when the alphabet consists of a large number of classes that share similar features. Such datasets are dominated by East Asian scripts, which typically include Chinese characters (Hanzi), Korean Hangul, and Japanese scripts (Kanji, which are derived from Chinese characters, as well as Hiragana and Katakana). To achieve higher accuracy, some large research teams use substantial computational resources to create deep models by overfitting the specific dataset. However, the same model may still perform unsatis-
factorily on the other dataset because it is not generalizable. More recent approaches, such as[10,11] focus on building lightweight offline classifiers that are easier to deploy, but they are incapable of demonstrating the same accuracy performance as some of the deeper state-of-the-art models.

In this paper, we present a scalable and comprehensive approach suitable for detailed character classification and show its efficiency on a highly complex handwritten character classification dataset CASIA-HWDB[12] that contains around 4 million images belonging to 7356 unique classes.

We also provide code and explanations and make an effort to create it suitable for the industry.

**2 Related Works**

Current works in HCR mainly focus on accuracy improvements and model complexity reductions.

The benchmark-level performance improvement is usually tackled by applying transfer learning,[13] increasing training dataset variation,[14] or analyzing handwriting styles.[15] Although the transfer learning technique gives notable accuracy improvements in comparison with lighter models, the effect of relying on this method can be neglected after passing a specific accuracy benchmark. Dataset variation increase can be achieved by augmentation, style transfer, cross-domain synthesis, or the leverage of generative adversarial networks for dataset enlargement. Handling the variation is a crucial step in the context of HCR. Real-world character images, especially the ones captured from books, introduce strong class and style imbalance,[16] making it complicated for the researchers to achieve high-level model generalization and scalability. The common preprocessing step used by the research teams is style analysis.[15,17] X. Li, J. Wang, H. Zhang, Y. Huang, and H. Huang in[18] demonstrated that the capability to detect and recognize font styles accurately has essential applications in graphic design, page layout, handwriting identification, and other use cases needed for building a solid intuition about the preprocessing steps for a successful classification algorithm.

Model complexity reduction is usually achieved through pruning algorithms and quantization, allowing lighter model usage without notable accuracy decrease. Through dynamic network pruning on SqueezeNet, an accuracy of 96.32% was achieved on the ICDAR-2013 dataset in.[19] By reconstructing SqueezeNext, introducing a cross-stage attention model and a new criterion for assessing the importance of convolutional kernels, Ruiqi Wu, Feng Zhou, Nan Li, Xian Liu, and Rugang Wang in[20] achieved 97.36% TOP-1 accuracy on CASIA-HWDB dataset. Chen et al. in[21] chose MobileNetV3 as a basis for using multiscale convolutional kernels and also received high accuracy by keeping the number of parameters low.

Deep convolutional neural networks (deep CNNs) can achieve high-level performance on complex datasets; however, using very deep CNNs is frequently not preferred due to a large number of parameters, leading to more extended training required, model storage complications, and the need to address learning trade-offs. Still, the deep CNN-based algorithms can be more straightforward and generalizable, making it possible to achieve exceptional performance on various datasets. In,[22] C. E. Mook, C. Poo Lee, K. M. Lim, and J. Yan Lim showed that the InceptionResNetV2 model with data augmentation achieved the highest accuracy across used benchmark datasets, with accuracies of 93.26%, 97.16%, and 99.71% on the English Handwritten Characters, Handwritten Digits, and MNIST datasets, respectively. It is also possible to introduce prediction generalization by ensembling the Deep CNN models. For example, the ZFNet, LeNet, AlexNet, VGG5, VGG-7, VGG-9, and VGG-16 prediction combination, followed by the SVM prediction layer, in[23] achieved 97.8% TOP-1 accuracy on the ICDAR-2013 dataset. To avoid model overfitting due to quick parameter saturation in Deep CNNs, H.-R Xiang et al. showed the classical deep ResNets' high-level performance on the CASIA-HWDB dataset.[24] The following techniques help deeper models to extract complex features without facing overfitting quickly when working with complex datasets. At the same time, they are highly scalable in comparison to predecessor model architectures. As research teams successfully construct more generalizable models, they encounter specific challenges in handling real-world data, including class imbalance, background complexity, and intra-class variation. These challenges require focused attention to ensure effective addressing of each complication. In our studies, we propose an approach that is capable of overcoming discussed problems.

**3 Method**

Due to its scalability and generalization, our approach can achieve highly competitive results. We divided the ideas discussed into two major groups:

1. Network Design

2. Predictive Design.

*3.1 Network Design*

We will illustrate the network design through Model Design 3.1.1, Loss Function 3.1.2, and Data Preprocessing 3.1.3. We use three independently trained models with similar architecture but different data preprocessing techniques to ensemble the resultant received weights.

*3.1.1 Model Design*

Each model is an ordered combination of learning bricks. Learning bricks consist of the following building blocks: convolutional blocks (Conv block), residual blocks (Res block), and inception blocks (Inception block). The learning brick-level model architecture visualization is demonstrated in Fig 1 We use the structured representation of the logic to highlight the importance of model scaling for a deep neural network design.

Having a total of five learning bricks, the model becomes a deep CNN-based model with a high level of scalability and generalization. The first four bricks share a similar structure but inhibit different hyperparameters. The auxiliary output (Aux output) follows each brick, and the main output (Main output) follows the last brick. The (1 - 4) brick building block-level architecture is demonstrated in Fig. 2.

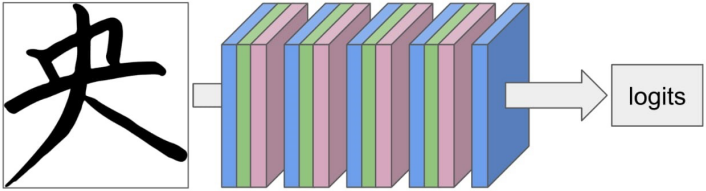

Fig 1 Learning brick-level Model Architecture.

The learning brick is represented by a sequential order of a convolutional block (Conv block), a residual block (Res block), and an inception block (Inception block).

Each convolutional block consists of four key components. First, a 2D convolutional layer is used to learn visual patterns and textures within the images. Through the application of trainable filters, this layer is able to capture spatial hierarchies of features at different levels of abstraction, from basic edges to more complex textures and patterns. Following the convolutional layer is a batch normalization layer. This layer helps to stabilize the learning process and accelerate the training procedure. It standardizes the input received by each unit in a mini-batch, which helps to ensure that inputs have a consistent mean and variance. This allows the network to learn more robustly using higher learning rates. Next, a 2D maximum pooling layer is used to reduce the spatial dimensions of the output volume from the previous layer. By applying a max pooling operation, the layer aggregates feature maps within local regions. This results in downsampling that makes the feature representations more tolerant to minor variations in the position of visual elements within the input images. Finally, a dropout layer is added as a regularization technique. During training, dropout randomly sets a fraction of the input units to zero at each update. This discourages complex co-adaptations on the training data and helps prevent overfitting. By ran-domly dropping out nodes and their connections, dropout forces the network to learn more robust and generalizable feature representations rather than focusing on memorizing intricate patterns in the training samples. Together, these four components allow each convolutional block to extract discriminative spatial features progressively.

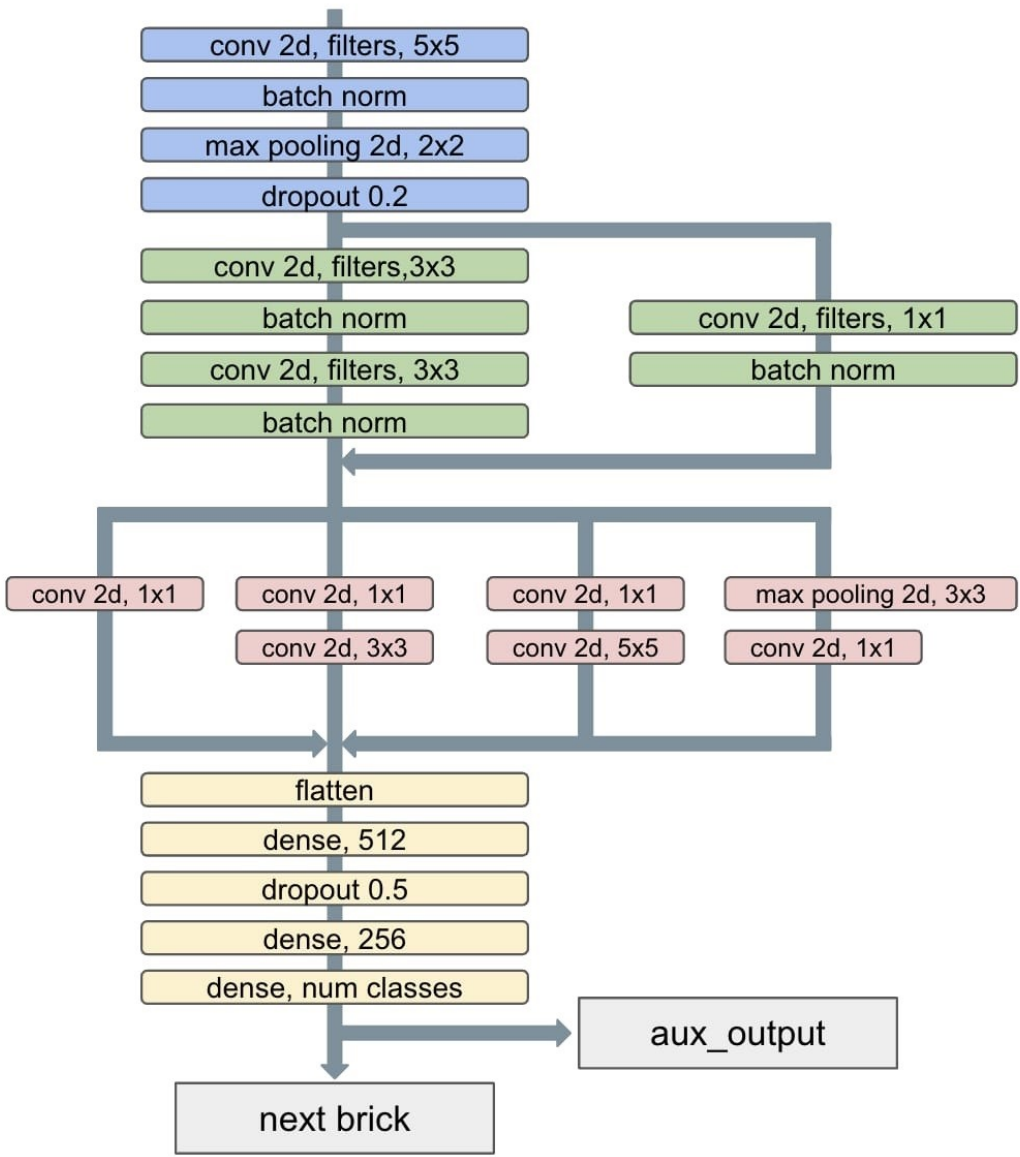

Fig 2 The 1st – 4th Brick Building block-level Architecture.

Each residual block[2] implementation contains a shortcut path and a main path. The shortcut path applies a 1x1 convolution with a stride of 2 to match the dimensions to the main path, followed by batch normalization. The main path first applies a regular 3x3 convolution with a stride of 2 for downsampling, followed by batch normalization and an activation function. It then applies another 3x3 convolution before batch normalization. The shortcut and main path outputs are summed. Another activation follows the addition to complete the main path. Finally, the residual block returns the combined output of the shortcut and main paths, representing the feature map resulting from the residual operation. The residual design is introduced to address the vanishing gradient problem that can occur in very deep convolutional networks. As the gradients are backpropagated, they can grow exponentially smaller with each layer until they essentially disappear. Residual blocks help alleviate this issue by introducing an additional pathway that skips over the intermediate layers. By adding a direct connection from input to output, residual blocks enable gradient signals to flow more smoothly across many layers during training. By leveraging skip connections, residual blocks make it easier for a network to learn complex functions by stacking representations. Overall, the residual blocks help drive performance improvements as networks scale to greater depths.

Each inception block is based on the inception module first introduced in GoogLeNet for largescale image classification.[25] The inception block is used to capture multi-scale information from its input more effectively. Concretely, an inception block applies multiple convolutional layers with different filter sizes: 1x1, 3x3, and 5x5 - in parallel to process the same input feature map. By utilizing filters of varying dimensions, the network can learn features at different levels of abstraction, from fine to coarse scales, within a single block. The outputs from the parallel convolutional layers are then concatenated, bringing together the diverse set of representations extracted at various scales. By evaluating multiple resolutions simultaneously and aggregating them, the inception block helps the model generalize better across spatial hierarchies present in the input data. It allows relevant patterns to emerge regardless of size, orientation, or precise location. This approach of performing mixed-size convolutions in parallel and concatenating the results significantly improves performance in recognition tasks compared to stacks of units operating on the entire image at a time.

Each brick inhibits a specific set of hyperparameters. The convolutional layers of the first brick use 32 filters, and the number of filters doubles in each next learning brick; this idea also applies to the inception block and residual block convolutional layers. The pooling size is always (2,2) to avoid rapid and unstable dimension reduction. The dropout inside each brick constantly equals 0.2; the more significant dropout is used in each output.

Each learning brick is followed by an auxiliary output branch aimed at guiding the training process. The auxiliary output first passes the extracted feature maps through a flattening layer, converting the spatial dimensions into a single vector. This flattened output is then fed into a series of densely connected layers. The dense layers allow the auxiliary branch to learn nonlinear interactions across the flattened features. A dropout layer is also included after the dense layers for regularization. By design, the auxiliary output branch helps to tackle the vanishing gradient problem during backpropagation. It provides additional gradient signals from the output that are fed back into earlier intermediate layers. The gradient feedback mechanism helps optimize the earlier representations learned by those layers. Additionally, utilizing auxiliary outputs at different depths in the model actively encourages the network to learn meaningful representations and abstractions as it progresses from early to later bricks. It reinforces the learning of valuable features at varying levels of abstraction. The loss calculated from each auxiliary branch gets applied to optimize the overall training process. As discussed in section 3.1.2, this auxiliary outputs approach facilitates faster convergence and improved performance of very deep networks compared to solely relying on a single output head.

The last brick (5) contains only a convolutional block. It has no maximum pooling layer but uses a stronger dropout. It is followed by a main output, used later for model evaluation. The main output includes a flattening layer, followed by a couple of dense layers with the strongest dropout that all together lead to a layer responsible for classification. The (5) brick building block-level architecture is demonstrated in Fig. 3.

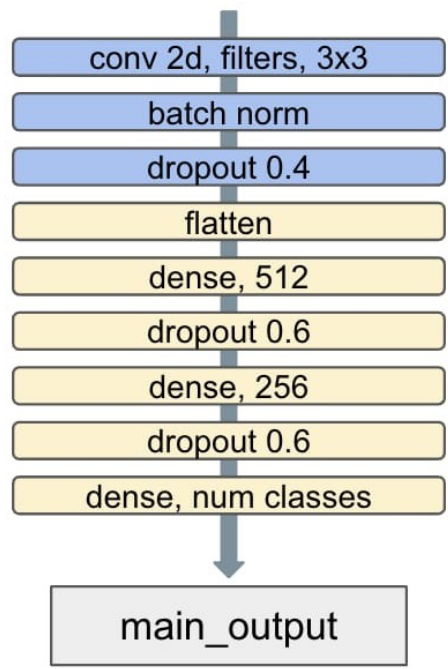

Fig 3 The 5th Brick Building block-level Architecture

The architecture design enables the model to learn both low-level and high-level features without losing the critical idea of modular scalability and generalization.

*3.1.2 Loss Function*

The models are trained using *α*-balanced focal cross-entropy loss. The focal loss function approach, introduced by Lin et al. in,[26] is implemented to address the class imbalance problem, which is a common challenge for the majority of real-world datasets. Oftentimes, imbalanced data prevents rare categories from contributing to the learning process. The *α*-balanced focal cross-entropy is derived from The *α*-balanced cross-entropy loss (Equation 1) by introducing a weighting factor *γ*.

$$CE(p,y) = CE(p_t) = \alpha_t \log(p_t) \qquad (1)$$

This factor *γ* controls each label's contribution to the overall loss function by assigning different significance levels for each class (Equation 2). The approach helps rare categories to contribute to the learning process by having a more considerable weight significance.

$$FCE(p_t) = -\alpha_t(1 - p_t)^\gamma \log(p_t) \qquad (2)$$

We suggest that the larger the *γ*-value is chosen, the longer the training should be performed. For our work, we chose the *γ*-value equal to 4, so that often-encountered classes would also have a considerable influence on the learning process. Experiment results are shown in section 4 of this paper.

Auxiliary outputs, together with the main output, contribute to the loss computation due to the loss function weight reassignment. The main output has a weight equal to 1, and the auxiliary outputs have assigned weights equal to 0.025, 0.05, 0.5, and 0.2 correspondingly. This helps the model to learn low-level features and, at the same time, extract high-level information from the images. The 3rd auxiliary output has a weight of 0.5 to stimulate mid-depth level feature exploration. The loss function weight reassignment may also be seen as a regularization technique. The combination of weights described demonstrates excellent performance; however, we encourage you to explore other potentially efficient weight sets.

*3.1.3 Data Preprocessing*

Model ensemble techniques such as averaging the predictions from multiple models trained on the same data have proven effective for improving generalization in competitive machine learning problems.[27,28] However, for CNNs designed for visual tasks, a more straightforward approach is to leverage data augmentation to enforce stronger generalization during training. In this work, we opt to train multiple instances of our CNN-based architecture on variants of the input data preprocessed in different ways, analogous to ensemble methods but operating at the data rather than model level.

Three distinct variants of the training dataset are prepared to train separate weight configurations of the model. The first utilizes the raw input images resized to 256x256 pixels and converted to grayscale. For the second variant, we apply Gaussian blurring with a kernel size of 3x3 and a standard deviation of 20 to each image before feeding to the model. Similarly, the third variant utilizes Gaussian blurring with a larger kernel of 7x7 and a standard deviation of 50. By introducing controlled perturbations to the inputs through blurring, the models are encouraged to learn features that are invariant to small deformations in the visual signal.

    The technique described is crucial for addressing both the gradient vanishing and the background complexity problems. The blur introduced allows the model to understand more comprehensive features without relying on image-specific or author-specific biases. At the same time, the background contrast complexity is significantly reduced, and the results of transformations performed are demonstrated in Fig. 4, Fig. 5, and Fig. 6, respectively.

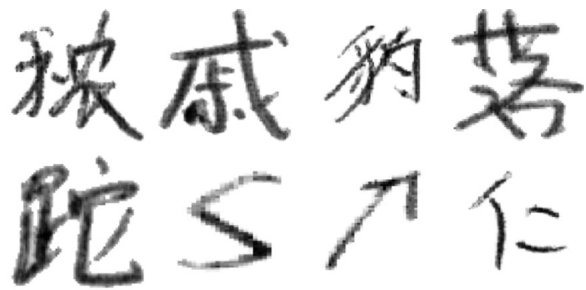

Fig 4 Image sample for Model 1 training, no Gaussian Blur.

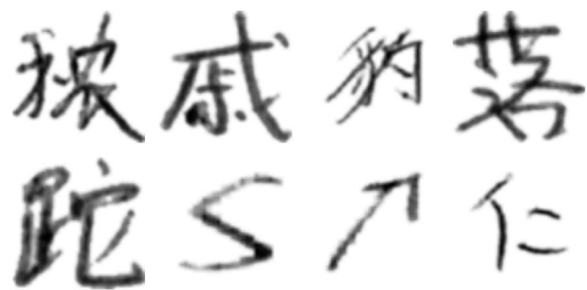

Fig 5 Image sample for Model 2 training, Gaussian Blur (3x3, 20).

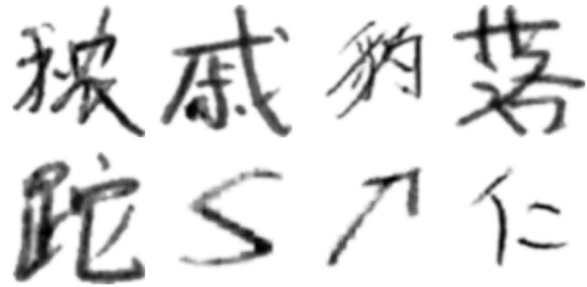

Fig 6 Image sample for Model 3 training, Gaussian Blur (5x5, 50).

*3.2 Predictive Design*

The model architecture leverages a weighted ensemble of independently trained models along with a multi-crop inference strategy to maximize predictive accuracy. Three separate model instances are trained on variants of the input data preprocessed using different Gaussian blur intensities. Their resulting weights are then optimally combined based on a weighted average scheme. Specifically, the first model trained on unchanged images is assigned a weight of 0.3. This reflects that features extracted from the raw inputs are essential; however, more abstract representations learning from blurred images help to generalize better. The second model using (3x3, 20) blur contributes with a weight of 0.2, emphasizing mid-level distortions. The third model's (5x5, 50) blur encourages highly invariant representations, thus receiving the highest weight of 0.5. The assigned weights were determined through cross-validation to achieve the best overall balanced performance. While these specific values work well, the weighting scheme is adaptable based on downstream task needs. The visualization is demonstrated in Fig. 7.

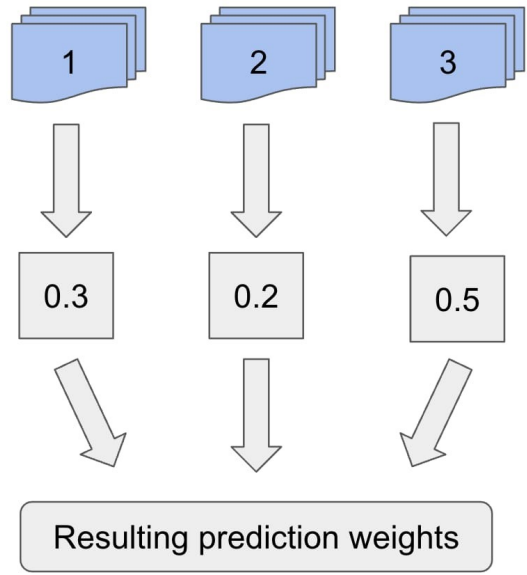

Fig 7 Proposed Model ensembling Weight Combination (3), (4), (5)- formulae.

At inference, a multi-view strategy further enhances predictions. Input images are first resized to 280x280 before extracting 5 cropped regions: 4 corners plus the center region at 256x256 resolution, same as training. The intermediate representations from each region are passed through the ensemble of models to generate class logits, which are averaged (Equation 3). Taking the softmax (Equation 4) and argmax (Equation 5) of this mean produces the final prediction. Compared to using a single model/crop, this multi-faceted approach considers an image's content more comprehensively through myriad context-preserving glimpses. The weighted ensemble and multi-crop testing protocol notably boosts result accuracy and consistency over baselines. However, further hyperparameter tuning may yield additional gains, with the optionality to explore more crop regions and weights. Overall, our approach demonstrates the benefits of principled, data-driven strategies for aggregating the learned knowledge. In section 4 of this paper, we perform experiments and show the results obtained.

$$L = \frac{1}{5} \sum_{5}^{i=1} L_i \tag{3}$$

$$P(c_k) = \frac{e^{L_k}}{\sum_j e^{L_j}} \tag{4}$$

$$PredictedClass = argmax_k P(c_k) \tag{5}$$

**4 Experiments**

We perform experiments on a CASIA-HWDB dataset that contains 3.9 million images of isolated characters that belong to 7356 classes produced by 1020 writers using Anoto pen on paper. The character images obtain high-level complexity and variation due to the way sampling was performed.[12] We shuffle the images and use 3 million for training and 0.9 million for testing. In Fig.8, we show the average accuracy of the three models trained depending on the γ-value chosen for the loss function.

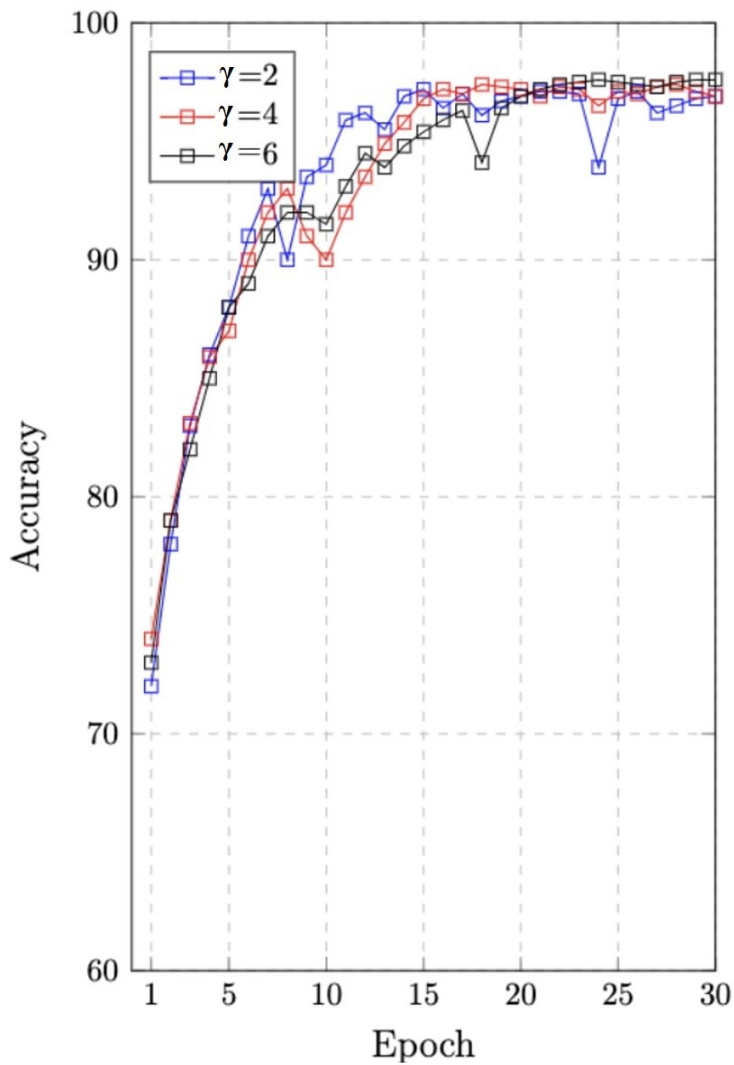

Fig 8 Accuracy Comparison for different $\gamma$-values.

It can be observed that the larger the $\gamma$-value is chosen, the higher the final accuracy is achieved. At the same time, learning takes more time because the algorithm downweights often-encountered classes and focuses on rare categories instead; this requires more time to find optimal weights. Better results can be achieved by paying more attention to rare classes using the loss function chosen. We suggest using $\gamma = 4$ to avoid more extended training and significant computational resource spending.

To evaluate our approach performance, we use some famous recent years famous models that have been assessed for the same classification task on the CASIA-HWDB dataset. We use HCCRGoogLeNet, Improved GoogLeNet, 2DPCA+ACNN, SqueezeNet+DNS, MSCS+ASA+SCL, HCCRMobileNetV3, LW-ViT, SqueezeNext+CCBAM as benchmarks, and the results of training with callbacks until no change for five epochs are demonstrated in Table 1. Although some models listed have advantages in size and parameter number, it is clear that our approach can learn highly complex features that lead to exceptional performance. We perform additional experiments to demonstrate the level of scalability, modality, generalization, and learning depth that the method can achieve and demonstrate the learning plateau the classical deep CNNs face.

Table 1 Model Comparison Performance Evaluation.

| Model | Year | Accuracy (%) |
|---|---|---|
| HCCR-GoogLeNet | 2019 | 96.31 |
| Improved GoogLeNet | 2020 | 97.48 |
| 2DPCA+ACNN | 2020 | 93.63 |
| SqueezeNet+DNS | 2021 | 96.32 |
| MSCS+ASA+SCL | 2022 | 97.63 |
| HCCR-MobileNetV3 | 2022 | 96.68 |
| LW-ViT | 2023 | 95.83 |
| SqueezeNext+CCBAM | 2023 | 97.36 |
| **Our Study** | 2024 | **97.79** |

We train the models for 50 epochs and compare the results achieved by AlexNet, HCCRGoogleNet, and our approach. The results of the accuracy achieved are shown in Table 2.

It is clear that AlexNet peaks at the lowest accuracy level and faces overfitting quickly. AlexNet can mainly leverage convolutions for the learning process, and the architecture has a limited number of regularization techniques integrated into the model design. Both HCCR-GoogleNet and our approach are deeper, more scalable, and more generalizable. They can extract a significant number of details and achieve better results which is crucial for a high complexity level dataset. At the same time, HCCR-GoogleNet and our approach do not hit clear overfitting; however, our methodology performs more stable and peaks at a higher accuracy level. The combination of experiments demonstrates that our approach, despite its comprehensiveness, has a strong advantage in feature learning. Being a deep CNN-based algorithm, the method is designed to address the common problems classical and modern deep CNNs face.

Table 2 Longer Training Model Comparison Performance Evaluation.

| Epoch Number | AlexNet | Improved GoogLeNet | **Our Study** |
|---|---|---|---|
| 5 | 81.94 | 84.22 | 84.77 |
| 10 | 85.76 | 89.98 | 89.52 |
| 15 | 90.99 | 94.11 | 93.34 |
| 20 | 91.54 | 96.78 | 96.84 |
| 25 | 92.02 | 97.29 | 97.59 |
| 30 | 91.51 | 97.02 | **97.72** |
| 35 | 90.20 | 96.72 | 97.39 |
| 40 | 85.46 | 96.89 | 96.91 |
| 45 | 86.78 | 94.99 | 97.02 |
| 50 | 83.89 | 96.73 | 97.48 |

To further develop the idea of the importance of scalability, modality, and generalization in the approach, we train for 30 epochs the 2-brick approach design, 3-brick approach design, 4- 4-brick approach design, and 5-brick approach design with γ = 4 for the loss function, the results are shown in Fig. 9.

It can be observed that deeper method models achieve better results due to the higher complexity features they can extract. Deeper models also demonstrate more stable performance. We show the easiness of scalability of the approach that makes it comprehensive.

The technique presented provides a flexible framework that makes the model construction pro-cess clear and organized. As dictated by modern theories around deep CNN scalability and evolving trends in the field, it is paramount to implement novel architectural components on top of established models in a way that demonstrates enhanced performance while preserving standard practices for generalization ability. Our approach meets this need by enriching robust baseline model ideas encountered in state-of-art models like GoogLeNet and ResNet through the application of well-known modules like inception blocks and residual connections and applying vital generalization techniques in a principled manner. Rather than designing radical new architectures from scratch, which risks losing generalization, we build upon foundations laid by previous works.

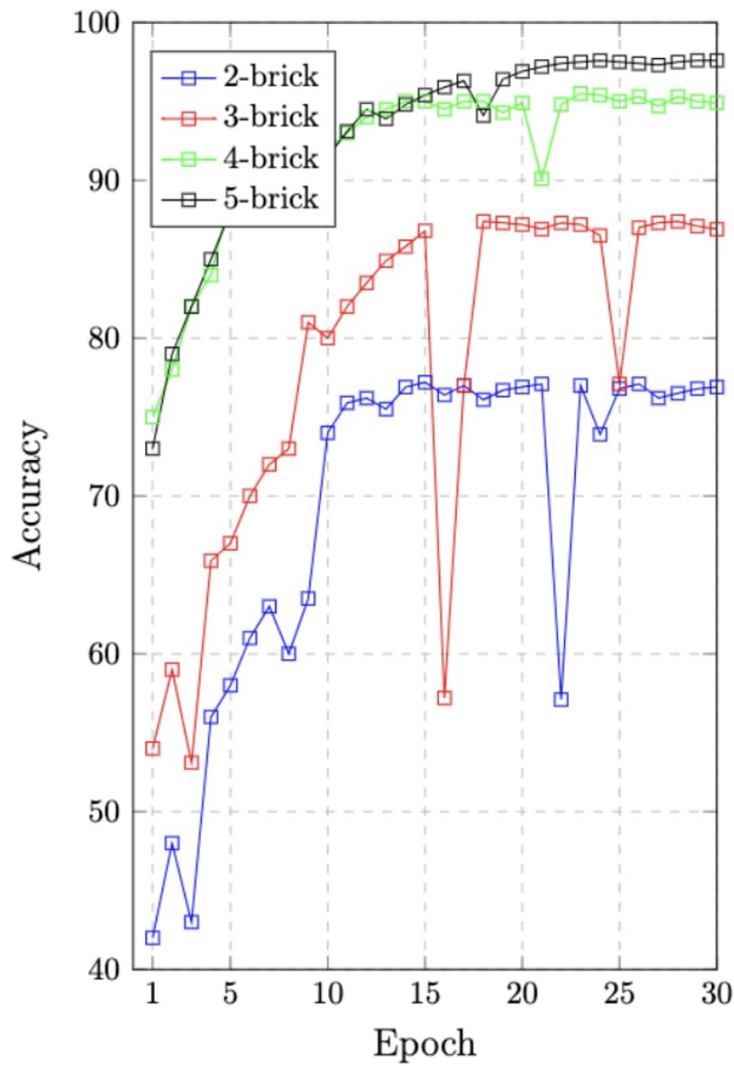

Fig 9 Brick-number comparison performance evaluation.

This allows new ideas to be transparently evaluated against solid baselines under controlled conditions. By working within the realm of established models, our method also enables easy technical replication and aids theoretical understanding of what components truly contribute to gains. Collectively, these characteristics make our technique uniquely suitable for transparently showcasing new modeling strategies through a process of refinement and verification, thus advancing both applied performance and scientific knowledge of deep CNN design.

Going forward, the flexibility of our platform implies it can seamlessly incorporate other promising modules or future works to distill technical progress into established frameworks. This facilitates ongoing progressive improvement following best practices and generalizability standards set forth in deep learning research.

## 5  Conclusion

In this work, we aim to comprehensively analyze the critical challenges facing handwritten character recognition (HCR) and identify the key limitations of many methodologies commonly utilized by researchers today. We state that scalability plays a crucial role in the model's ability to extract highly complex features without losing generalization. To tackle these ongoing issues, we propose a generalized deep learning-based approach that can be applied to any HCR task without extensive reworking or loss of generalization capabilities. The main strength of our proposed method lies in its simplicity, modularity, replicability, and ability to handle a wide variety of HCR problems in a scalable manner. Our methodology introduces novel yet intuitive techniques that encourage the learning of more robust representations without sacrificing standard best practices. To validate the efficacy of our idea, we conduct extensive experimental evaluations using the well-established CASIA-HWDB dataset, one of the most widely used benchmarks for offline handwritten Chinese character recognition. The results demonstrate that our approach is capable of achieving state-ofthe-art accuracy levels on this dataset, outperforming traditional methods. In summary, through our rigorous experiments and comparative studies, we confirm that the proposed approach establishes a new standard "flagship" solution for HCR tasks that can achieve high recognition accuracy without compromising scalability or flexibility of application to other domains and datasets.

## 6  Discussion

We believe that the introduced method can be used as a basis for further development and studies. The approach combines multiple highly effective ideas applicable to general usage. Each part can be isolated for further improvement without losing the generality of the other specific parts. The solution can become an effective platform for introducing ideas in the community and demonstrating the influence of new solutions by the evaluation metrics used.

*Disclosures*

All the authors declare no conflicts of interest.

*Acknowledgments*

This work is done with the guidance of Sparcus Technologies Limited, a leading research-oriented machine learning and data science agency with headquarters in Hong Kong, HKSAR.

*Code, Data, and Materials Availability*

The code is uploaded to GitHub and is available at: https://github.com/sparcus-technologies/ Deep-Learning-Driven-Approach-for-Handwritten-Chinese-Character-Classificat tree/main The dataset is provided by the Chinese Academy of Sciences and is available at: https://nlpr.ia.ac.cn/databases/handwriting/Home.html.

Boris Kriuk is the Co-Founder and CEO at Sparcus Technologies Limited, Machine Learning Lead at Antei Limited, LinkedIn Top Machine Learning Voice. His research interests lie in the field of deep learning in computer vision. Boris Kriuk is the co-author of two conference papers. Boris Kriuk is an undergraduate student at the Hong Kong University of Science and Technology, pursuing a degree in computer engineering.

Fedor Kriuk is the Co-Founder and CTO at Sparcus Technologies Limited. His research interests include statistical machine learning and deep learning in computer vision. Fedor Kriuk is the co-author of two conference papers. Fedor Kriuk is an undergraduate student at the Hong Kong University of Science and Technology, pursuing a degree in electrical engineering.

## List of Figures

## List of Tables